\documentclass{article}
\usepackage{amsmath,epsfig}

\usepackage{graphicx} 
\usepackage{subfigure} 
\usepackage{paralist}
\usepackage{MnSymbol}
\usepackage{natbib}

\usepackage{algorithm}
\usepackage{algorithmic}

\usepackage{hyperref}

\usepackage[accepted]{icml2014}

\icmltitlerunning{\emph{k}-Sparse Autoencoders}

\begin{document} 

\twocolumn[
\icmltitle{\emph{k}-Sparse Autoencoders}

\icmlauthor{Alireza Makhzani}{makhzani@psi.utoronto.ca}
\icmlauthor{Brendan Frey}{frey@psi.utoronto.ca}
\icmladdress{University of Toronto, 10 King's College Rd. Toronto, Ontario M5S 3G4, Canada}

\vskip 0.3in
]

\begin{abstract}
Recently, it has been observed that when representations are learnt in a way that encourages sparsity, improved performance is obtained on classification tasks. These methods involve combinations of activation functions, sampling steps and different kinds of penalties. To investigate the effectiveness of sparsity by itself, we propose the ``\emph{k}-sparse autoencoder'', which is an autoencoder with linear activation function, where in hidden layers only the $k$ highest activities are kept. When applied to the MNIST and NORB datasets, we find that this method achieves better classification results than denoising autoencoders, networks trained with dropout, and RBMs. \emph{k}-sparse autoencoders are simple to train and the encoding stage is very fast, making them well-suited to large problem sizes, where conventional sparse coding algorithms cannot be applied.
\end{abstract} 

\section{Introduction}

Sparse feature learning algorithms range from sparse coding approaches \cite{sparse_coding} to training neural networks with sparsity penalties \cite{lifetime,ng}. These methods typically comprise two steps: a learning algorithm that produces a dictionary $W$ that sparsely represents the data $\{\boldsymbol{x}_i\}_{i=1}^N$, and an encoding algorithm that, given the dictionary, defines a mapping from a new input vector $\boldsymbol{x}$ to a feature vector. 

A practical problem with sparse coding is that both the dictionary learning and the sparse encoding steps are computationally expensive. Dictionaries are usually learnt offline by iteratively recovering sparse codes and updating the dictionary. Sparse codes are computed using the current dictionary $W$ and a pursuit algorithm to solve
\begin{equation}
\hat{\boldsymbol{z}}_i=\underset{\boldsymbol{z}}{\text{argmin}} \|\boldsymbol{x}_i-W \boldsymbol{z}\|_2^2  \hspace{.25cm} s.t. \hspace{.25cm} \|\boldsymbol{z}\|_0 < \text{\emph{k}}
\end{equation}
where $\boldsymbol{z}_i$, $i=1,..,N$ are the columns of $Z$. Convex relaxation methods such as $\ell_1$ minimization or greedy methods such as OMP \cite{omp} are used to solve the above optimization. While greedy algorithms are faster, they are still slow in practice. The current sparse codes are then used to update the dictionary, using techniques such as the method of optimal directions (MOD) \cite{mod} or K-SVD \cite{ksvd}. These methods are computationally expensive; MOD requires inverting the data matrix at each step and K-SVD needs to compute a SVD in order to update every column of the dictionary.

To achieve speedups, in \cite{fast,psd}, a parameterized non-linear encoder function is trained to explicitly predict sparse codes using a soft thresholding operator. However, they assume that the dictionary is already given and do not address the offline phase.

Another approach that has been taken recently is to train autoencoders in a way that encourages sparsity. However, these methods usually involve combinations of activation functions, sampling steps and different kinds of penalties, and are sometimes not guaranteed to produce sparse representations for each input. For example, in \cite{ng,lifetime}, a ``lifetime sparsity'' penalty function proportional to the negative of the KL divergence between the hidden unit marginals and the target sparsity probability is added to the cost function. This results in sparse activation of hidden units across training points, but does not guarantee that each input has a sparse representation.

\begin{inparaenum}[(i)] The contributions of this paper are as follows. \item We describe ``\emph{k}-sparse autoencoders" and show that they can be efficiently learnt and used for sparse coding. \item We explore how different sparsity levels ($k$) impact representations and classification performance. \item We show that by solely relying on sparsity as the regularizer and as the \emph{only nonlinearity}, we can achieve much better results than the other methods, including RBMs, denoising autoencoders \cite{da} and dropout \cite{dropout}. \item We demonstrate that \emph{k}-sparse autoencoders are suitable for pretraining and achieve results comparable to state-of-the-art on MNIST and NORB datasets. \end{inparaenum}

In this paper, $\Gamma$ is an estimated support set and $\Gamma^c$ is its complement. $W^\dagger$ is the pseudo-inverse of $W$ and $\text{supp}_\text{\emph{k}}(\boldsymbol{x})$ is an operator that returns the indices of the \emph{k} largest coefficients of its input vector. $\boldsymbol{z}_\Gamma$ is the vector obtained by restricting the elements of  $\boldsymbol{z}$ to the indices of $\Gamma$ and $W_\Gamma$ is the matrix obtained by restricting the columns of $W$ to the indices of $\Gamma$.

\section{Description of the Algorithm} 

\subsection{The Basic Autoencoder} 
\label{sub:the_basic_}
A shallow autoencoder maps an input vector $\boldsymbol{x}$ to a hidden representation using the function $\boldsymbol{z}=f(P \boldsymbol{x} + \boldsymbol{b})$, parameterized by $\{P,\boldsymbol{b}\}$. $f$ is the activation function, \emph{e.g.}, linear, sigmoidal or ReLU. The hidden representation is then mapped linearly to the output using $\hat{\boldsymbol{x}}=W \boldsymbol{z} +\boldsymbol{b}'$. The parameters are optimized to minimize the mean square error of $\| \hat{\boldsymbol{x}}-\boldsymbol{x}\|_2^2$ over all training points. Often, tied weights are used, so that $P=W^\top$.

\subsection{The \emph{k}-Sparse Autoencoder} 
\label{k_sparse}

The $k$-sparse autoencoder is based on an autoencoder with linear activation functions and tied weights. In the feedforward phase, after computing the hidden code $\boldsymbol{z}=W^\top  \boldsymbol{x} + \boldsymbol{b}$, rather than reconstructing the input from all of the hidden units, we identify the \emph{k} largest hidden units and set the others to zero. This can be done by sorting the activities or by using ReLU hidden units with thresholds that are adaptively adjusted until the $k$ larges activities are identified. This results in a vector of activities with the support set of $\text{supp}_\text{\emph{k}}(W^\top  \boldsymbol{x}+\boldsymbol{b})$. Note that once the \emph{k} largest activities are selected, the function computed by the network is linear. So the only non-linearity comes from the selection of the \emph{k} largest activities. This selection step acts as a regularizer that prevents the use of an overly large number of hidden units when reconstructing the input.

Once the weights are trained, the resulting sparse representations may be used for learning to perform downstream classification tasks. However, it has been observed that often, better results are obtained when the sparse encoding stage used for classification does not exactly match the encoding used for dictionary training \cite{adam}. For example, while in \emph{k}-means, it is natural to have a hard-assignment of the points to the nearest cluster in the encoding stage, it has been shown in \cite{mix} that soft assignments  result in better classification performance. Similarly, for the \emph{k}-sparse autoencoder, instead of using the \emph{k} largest elements of $W^\top  \boldsymbol{x} + \boldsymbol{b}$ as the features, we have observed that slightly better performance is obtained by using the \emph{$\alpha$k} largest hidden units where $\alpha \geq 1$ is selected using validation data. So at the test time, we use the support set defined by $\text{supp}_{\alpha\emph{k}}(W^\top  \boldsymbol{x} + \boldsymbol{b})$. The algorithm is summarized as follows.

\begin{center}
\begin{tabular}{|l|}
  \hline
  \emph{k}-Sparse Autoencoders:  \\
  \hline

\textbf{Training:}\\

\hspace{.2cm}1) Perform the feedforward phase and compute \\
\hspace{3cm}$\boldsymbol{z} = W^\top  \boldsymbol{x} + \boldsymbol{b}$\\
\hspace{.2cm}2) Find the \emph{k} largest activations of $\boldsymbol{z}$ and set\\ 
\hspace{.2cm}the rest to zero.\\
\hspace{1.5cm} $\boldsymbol{z}_{(\Gamma)^c}=0$ \hspace{.1cm} where  \hspace{.1cm} $\Gamma = \text{supp}_{\text{\emph{k}}} (\boldsymbol{z}) $\\
\hspace{.2cm}3) Compute the output and the error using the\\
\hspace{.2cm}sparsified $\boldsymbol{z}$.\\
\hspace{3cm}$\hat{\boldsymbol{x}}=W \boldsymbol{z} +\boldsymbol{b}'$\\
\hspace{3cm}$E =\|\boldsymbol{x}-\hat{\boldsymbol{x}}\|_2^2$\\
\hspace{.2cm}3) Backpropagate the error through the \emph{k} largest\\ 
\hspace{.2cm}activations defined by $\Gamma$ and iterate.\\

\textbf{Sparse Encoding:}\\

\hspace{.2cm}Compute the features $\boldsymbol{h} = W^\top  \boldsymbol{x} + \boldsymbol{b}$. Find its $\alpha\text{\emph{k}}$\\
\hspace{.2cm}largest activations and set the rest to zero.\\
\hspace{1.5cm} $\boldsymbol{h}_{(\Gamma)^c}=0$ \hspace{.2cm} where  \hspace{.2cm} $\Gamma = \text{supp}_{\alpha\text{\emph{k}}} (\boldsymbol{h}) $\\
  \hline
\end{tabular}
\end{center}

\section{Analysis of the \emph{k}-Sparse Autoencoder} 
\label{analysis}

In this section, we explain how the \emph{k}-sparse autoencoder can be viewed in the context of sparse coding with incoherent matrices. This perspective sheds light on why the \emph{k}-sparse autoencoders work and why they achieve invariant features and consequently good classification results. We first explain a sparse recovery algorithm and then show that the \emph{k}-sparse autoencoder iterates between an approximation of this algorithm and a dictionary update stage.

\subsection{\small{Iterative Thresholding with Inversion (ITI)}} 
\label{sub:iti_algorithm}

Iterative hard thresholding \cite{iht} is a class of low complexity algorithms, which has recently been  proposed for the reconstruction of sparse signals. In this work, we use a variant called ``iterative thresholding with inversion" \cite{coherence}. Given a fixed $\boldsymbol{x}$ and $W$, starting from $\boldsymbol{z}^0=0$, ITI iteratively finds the sparsest solution of $\boldsymbol{x}=W \boldsymbol{z}$ using the following steps.

\begin{enumerate}
	\item \textbf{Support Estimation Step:}
\begin{equation}
\Gamma = \text{supp}_\text{\emph{k}}(\boldsymbol{z}^n+W^\top (\boldsymbol{x}-W \boldsymbol{z}^n))
\end{equation}
	\item \textbf{Inversion Step:}
\begin{equation}
\begin{aligned}
\boldsymbol{z}_{\Gamma}^{n+1} &= W_\Gamma^\dagger \boldsymbol{x} = \text{ }(W_{\Gamma}^\top W_{\Gamma})^{-1}W_{\Gamma}^\top \boldsymbol{x} \\
\boldsymbol{z}_{(\Gamma)^c}^{n+1} &= \text{ } 0
\end{aligned}
\end{equation}

\end{enumerate}

Assume $H= W^\top  W - I$ and $\boldsymbol{z}_0$ is the true sparse solution. The first step of ITI estimates the support set as $\Gamma=\text{supp}_\text{\emph{k}} (W^\top  \boldsymbol{x}) = \text{supp}_\text{\emph{k}} (\boldsymbol{z}_0 + H \boldsymbol{z}_0)$. If $W$ was orthogonal, we would have $H \boldsymbol{z_0} = 0$ and the algorithm would succeed in the first iteration. But if $W$ is overcomplete, $H \boldsymbol{z}_0$ behaves as a noise vector whose variance decreases after each iteration. After estimating the support set of $\boldsymbol{z}$ as $\Gamma$, we restrict $W$ to the indices included in $\Gamma$ and form $W_\Gamma$. We then use the pseudo-inverse of $W_\Gamma$ to estimate the non-zero values minimizing $\|\boldsymbol{x} - W_{\Gamma} \boldsymbol{z}_{\Gamma}\|_2^2$. Lastly, we refine the support estimation and repeat the whole process until convergence.

\subsection{Sparse Coding with the \emph{k}-Sparse Autoencoder} 
\label{sub:dictionary_learning}

Here, we show that we can derive the \emph{k}-sparse autoencoder tarining algorithm by approximating a sparse coding algorithm that uses the ITI algorithm jointly with a dictionary update stage.

The conventional approach of sparse coding is to fix the sparse code matrix $Z$, while updating the dictionary. However, here, after estimating the support set in the first step of the ITI algorithm, we jointly perform the inversion step of ITI and the dictionary update step, while fixing just the support set of the sparse code $Z$. In other words, we update the atoms of the dictionary and allow the corresponding non-zero values to change at the same time to minimize $\|X - W_{\Gamma} Z_{\Gamma}\|_2^2$ over both $W_{\Gamma}$ and $Z_{\Gamma}$.

When we are performing sparse recovery with the ITI algorithm using a fixed dictionary, we should perform a fixed number of iterations to get the perfect reconstruction of the signal. But, in sparse coding, since we learnt a dictionary that is adapted to the signals, as shown in Section \ref{incoherence}, we can find the support set just with the first iteration of ITI: 

\begin{equation}
	\Gamma_{\boldsymbol{z}}=\text{supp}_\text{\emph{k}}(W^\top  \boldsymbol{x})
\end{equation}

In the inversion step of the ITI algorithm, once we estimate the support set, we use the pseudo-inverse of $W_\Gamma$ to find the non-zero values of the support set. The pseudo-inverse of the matrix $W_\Gamma$ is a matrix, such as $P_\Gamma$, that minimizes the following cost function.

\begin{equation}
  \begin{aligned}
W_\Gamma^\dagger &= \underset{P_\Gamma}{\text{arg min}} \|\boldsymbol{x} -W_\Gamma \boldsymbol{z}_\Gamma\|_2^2 \\  
&= \underset{P_\Gamma}{\text{arg min}} \|\boldsymbol{x} -W_\Gamma P_\Gamma \boldsymbol{x} \|_2^2
  \end{aligned}
\end{equation}

Finding the exact pseudo-inverse of $W_\Gamma$ is computationally expensive, so instead, we perform a single step of gradient descent. The gradient with respect to $P_\Gamma$ is found as follows:

\begin{equation}
\label{eq:grad}
	{\partial \|\boldsymbol{x} -W_\Gamma \boldsymbol{z}_\Gamma \|_2^2 \over \partial P_\Gamma} = {\partial \|\boldsymbol{x} -W_\Gamma \boldsymbol{z}_\Gamma \|_2^2 \over \partial \boldsymbol{z}_\Gamma} \boldsymbol{x}
\end{equation}

The first term of the right hand side of the Equation \eqref{eq:grad} is the dictionary update stage, which is computed as follows:

\begin{equation}
	{\partial \|\boldsymbol{x} -W_\Gamma \boldsymbol{z}_\Gamma \|_2^2 \over \partial \boldsymbol{z}_\Gamma} = (W_\Gamma \boldsymbol{z}_\Gamma - \boldsymbol{x})  \boldsymbol{z}^\top _\Gamma
\end{equation}

Therefore, in order to approximate the pseudo-inverse, we first find the dictionary derivative and then ``backpropagate" it to find the update of the pseudo-inverse. 

We can view these operations in the context of an autoencoder with linear activations where $P$ is the encoder weight matrix and $W$ is the decoder weight matrix. At each iteration, instead of back-propagating through all the hidden units, we just back-propagate through the units with the \emph{k} largest activities, defined by $\text{supp}_\text{\emph{k}}(W^\top  \boldsymbol{x})$, which is the first iteration of ITI. Keeping the \emph{k} largest hidden activities and ignoring the others is the same as forming $W_\Gamma$ by restricting $W$ to the estimated support set. Back-propagation on the decoder weights is the same as gradient descent on the dictionary and back-propagation on the encoder weights is the same as approximating the pseudo-inverse of the corresponding $W_\Gamma$.

We can perform support estimation in the feedforward phase by assuming $P = W^\top $ (i.e., the autoencoder has tied weights). In this case, support estimation can be done by computing $\boldsymbol{z}=W^\top  \boldsymbol{x} + \boldsymbol{b}$ and picking the \emph{k} largest activations; the bias just accounts for the mean and subtracts its contribution. Then the ``inversion" and ``dictionary update" steps are done at the same time by back-propagation through just the units with the \emph{k} largest activities.

In summary, we can view \emph{k}-sparse autoencoders as the approximation of a sparse coding algorithm which uses ITI in the sparse recovery stage.

\subsection{Importance of Incoherence}
\label{incoherence}

The coherence of a dictionary indicates the degree of similarity between different atoms or different collections of atoms. Since the dictionary is overcomplete, we can represent each column of the dictionary as a linear combination of other columns. But what incoherence implies is that we should not be able to represent a column as a \emph{sparse} linear combination of other columns and the coefficients of the linear combination should be dense. For example, if two columns are exactly the same, then the dictionary is highly coherent since we can represent one of those columns as the sparse linear combination of the rest of the columns. 
A naive measure of coherence that has been proposed in the literature is the mutual coherence $\mu(W)$ which is defined as the maximum absolute inner product across all the possible pairs of the atoms of the dictionary.
\begin{equation}
\mu(W)= \underset{i \neq j}{\text{max }}  |\langle  \boldsymbol{w}_i, \boldsymbol{w}_j \rangle|
\end{equation}

There is a close relationship between the coherency of the dictionary and the uniqueness of the sparse solution of $\boldsymbol{x} = W \boldsymbol{z}$. In \cite{donoho}, it has been proven that if $\text{\emph{k}} \leq (1+\mu^{-1})$, then the sparsest solution is unique. 

We can show that if the dictionary is incoherent enough, there is going to be an attraction ball around the signal $\boldsymbol{x}$ and there is only one unique sparse linear combination of the columns that can get into this attraction ball. So even if we perturb the input with a small amount of noise, translation, rotation, \emph{etc.}, we can still achieve perfect reconstruction of the original signal and the sparse features are always roughly conserved. Therefore, incoherency of the dictionary is a measure of invariance and stability of the features. This is related to the denoising autoencoder \cite{da} in which we achieve invariant features by trying to reconstruct the original signal from its noisy versions.

Here we show that if the dictionary is incoherent enough, the first step of the ITI algorithm is sufficient for perfect sparse recovery.

\textbf{Theorem 3.1.} Assume $\boldsymbol{x} = W \boldsymbol{z}$ and the columns of the dictionary have unit $\ell_2$-norm. Also without loss of generality, assume that the non-zero elements of $\boldsymbol{z}$ are its first \emph{k} elements and are sorted as $z_1 \geq z_2 \geq ... \geq z_\text{\emph{k}}$.
Then, if ${\text{\emph{k}}\mu \leq {z_\text{\emph{k}} \over {2 z_1}}}$, we can recover the support set of $\boldsymbol{z}$ using $\text{supp}_\text{\emph{k}}(W^\top  \boldsymbol{x})$. 

\textbf{Proof}: Let us assume $0 \leq i \leq \text{\emph{k}}$ and $\boldsymbol{y}=W^\top  \boldsymbol{x}$. Then, we can write:

\begin{equation}
 \begin{aligned}
	y_i &= z_i+\sum_{j=1,j \neq i}^{\text{\emph{k}}} \langle \boldsymbol{w}_i,\boldsymbol{w}_j \rangle z_j 
	&\geq z_i - \mu \sum_{j=1,j \neq i}^{\text{\emph{k}}} z_j 
	&\geq z_\text{\emph{k}} - \text{\emph{k}} \mu z_1  
  \end{aligned}
\end{equation}
On the other hand:
\begin{equation}
 \begin{aligned}
  	\underset{i>\text{\emph{k}}}{\text{max }} \{y_i\} &= \underset{i>\text{\emph{k}}}{\text{max }} \left\{ \sum_{j=1}^{\text{\emph{k}}} \langle \boldsymbol{w}_i,\boldsymbol{w}_j \rangle z_j \right\}
  	&\leq \text{\emph{k}}\mu z_1  
  \end{aligned}
\end{equation}

So if $\text{\emph{k}}\mu \leq {z_\text{\emph{k}} \over {2 z_1}}$, all the first \emph{k} elements of $\boldsymbol{y}$ are guaranteed to be greater than the rest of its elements.

As we can see from Theorem 3.1, the chances of finding the true support set with the encoder part of the \emph{k}-sparse autoencoder depends on the incoherency of the learnt dictionary. As the \emph{k}-sparse autoencoder converges (i.e., the reconstruction error goes to zero), the algorithm learns a dictionary that satisfies $\boldsymbol{x} \approx W \boldsymbol{z}$, so the support set of $\boldsymbol{z}$ can be estimated using the first step of ITI. Since $\text{supp}_\text{\emph{k}}(W^\top  \boldsymbol{x})$ succeeds in finding the support set when the algorithm converges, the learnt dictionary must be sufficiently incoherent. 

\section{Experiments}

In this section, we evaluate the performance of \emph{k}-sparse autoencoders in both unsupervised learning and in shallow and deep discriminative learning tasks.

\subsection{Datasets}

We use the MNIST handwritten digit dataset, which consists of 60,000 training images and 10,000 test images. We randomly separate the training set into 50,000 training cases and 10,000 cases for validation.

We also use the small NORB normalized-uniform dataset \cite{norb}, which contains 24,300 training examples and 24,300 test examples. This database contains images of 50 toys from 5 generic categories: four-legged animals, human figures, airplanes, trucks, and cars. Each image consists of two channels, each of size $96 \times 96$ pixels. We take the inner $64 \times 64$ pixels of each channel and resize it using bicubic interpolation to the size of $32 \times 32$ pixels from which we form a vector with 2048 dimensions as the input. Data points are subtracted by the mean and divided by the standard deviation along each input dimension across the whole training set to normalize the contrast. The training set is separated into 20,000 for training and 4,300 for validation.

We also test our method on natural image patches extracted from CIFAR-10 dataset. We randomly extract 1000000 patches of size $8 \times 8$ from the 50000 $32 \times 32$ images of CIFAR-10. Each patch is then locally contrast-normalized and ZCA whitened. This preprocessing pipeline is the same as the one used in \cite{cifar10_adam} for feature extraction.

\begin{figure*}[h!tb]
\centering
\subfigure[$\text{\emph{k}}=70$]{
\includegraphics[scale=.25]{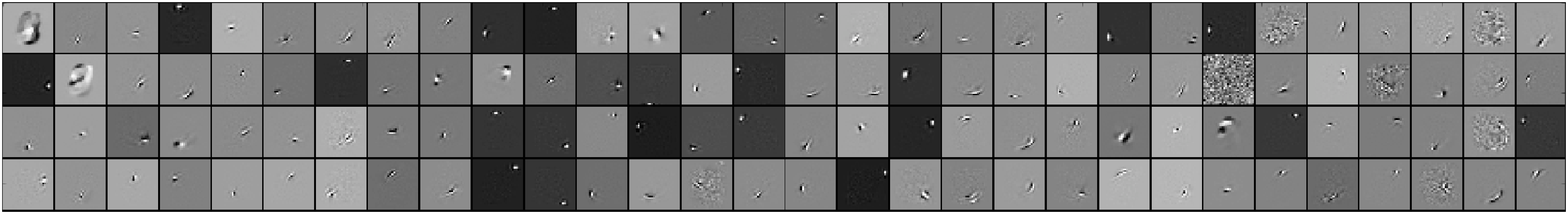}}
\subfigure[$\text{\emph{k}}=40$]{
\hspace{.2cm}\includegraphics[scale=.25]{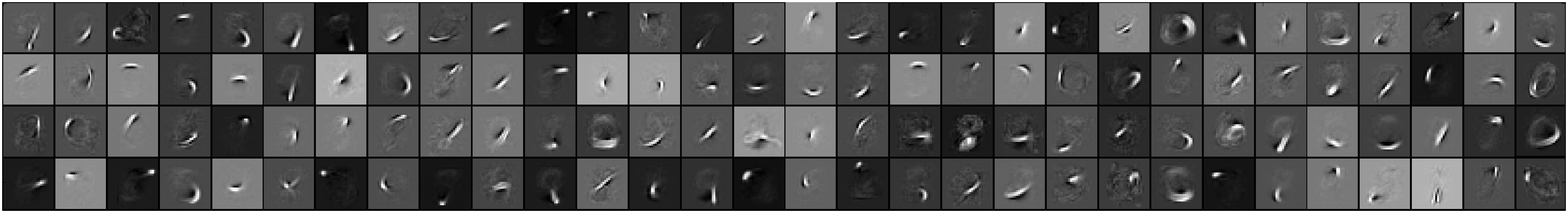}}
\subfigure[$\text{\emph{k}}=25$]{
\hspace{.2cm}\includegraphics[scale=.25]{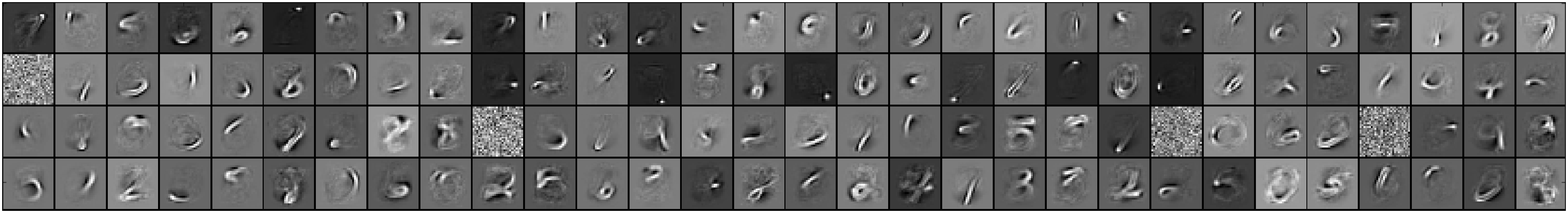}}
\subfigure[$\text{\emph{k}}=10$]{
\includegraphics[scale=.25]{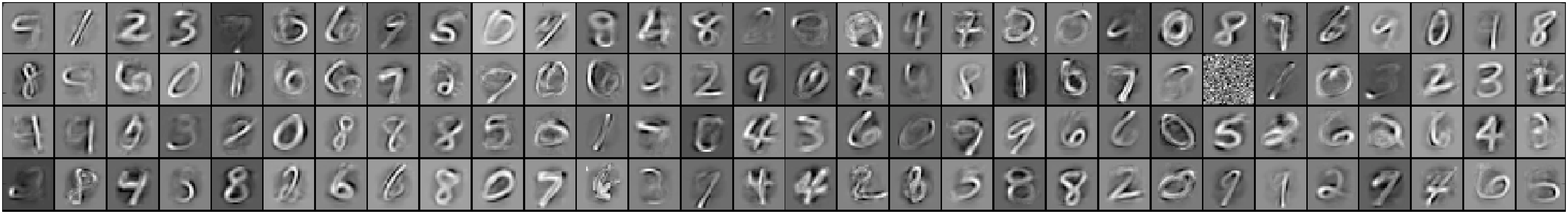}}
\caption{\label{fig_mnist}Filters of the \emph{k}-sparse autoencoder for different sparsity levels \emph{k}, learnt from MNIST with 1000 hidden units.}
\end{figure*}

\begin{figure*}[h!tb]
\centering
\subfigure[$\text{\emph{k}}=200$]{
\includegraphics[scale=.392]{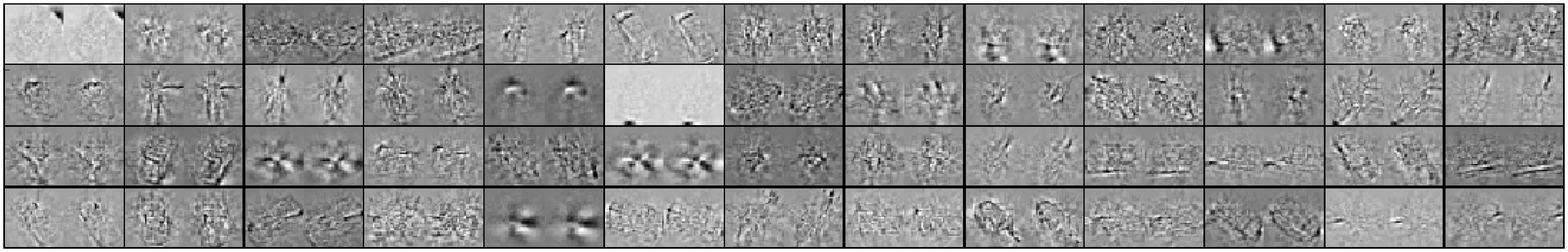}}
\subfigure[$\text{\emph{k}}=150$]{
\includegraphics[scale=.425]{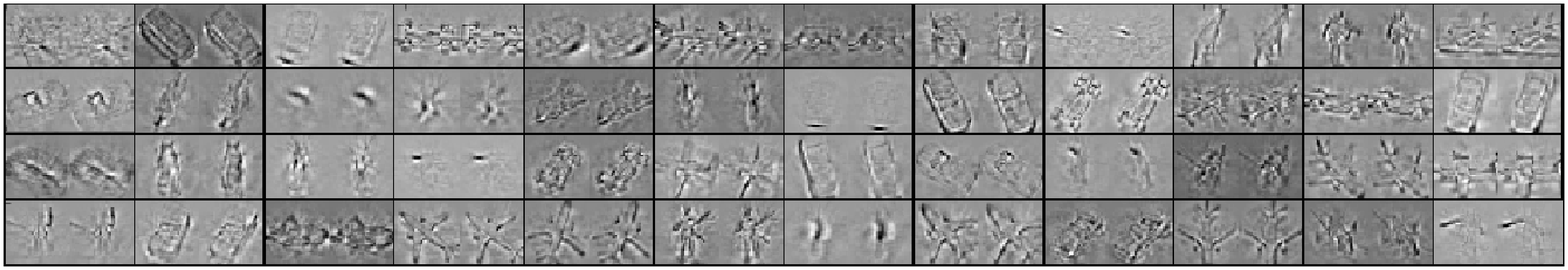}}
\subfigure[$\text{\emph{k}}=50$]{
\includegraphics[scale=.25]{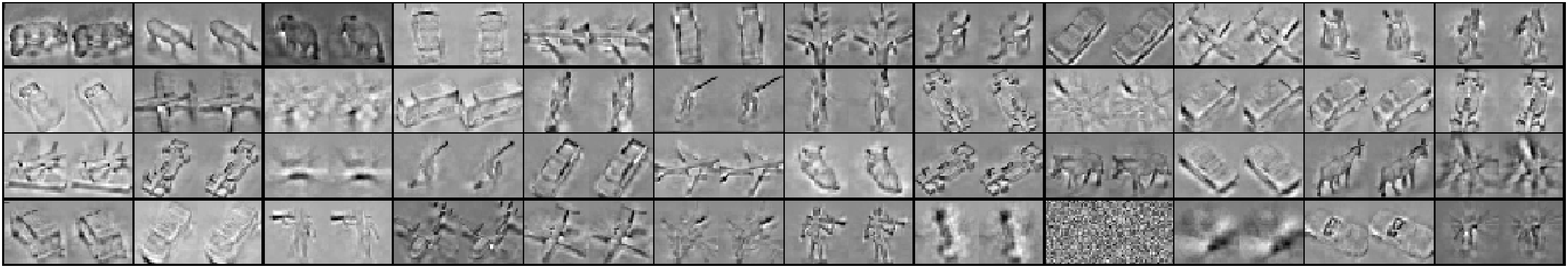}}
\caption{\label{fig_norb}Filters of the \emph{k}-sparse autoencoder for different sparsity levels \emph{k}, learnt from NORB with 4000 hidden units.}
\end{figure*}

\begin{table*}[h!tb]
\parbox{.50\linewidth}{
\centering 
\begin{tabular}{ |l  | r| }
  \hline
   & Error Rate \\
  \hline 
  Raw Pixels  & 7.20\% \\
  RBM  & 1.81\%\\
  Dropout Autoencoder \small{(50\% hidden)} & 1.80\% \\
  Denoising Autoencoder     & 1.95\% \\
  \small{(20\% input dropout)}&\\
  Dropout + Denoising Autoencoder  & 1.60\% \\
  \small(20\% input and 50\% hidden) & \\
  \emph{k}-Sparse Autoencoder, \emph{k} = 40 & 1.54\% \\
  \emph{k}-Sparse Autoencoder, \emph{k} = 25  & 1.35\% \\
  \emph{k}-Sparse Autoencoder, \emph{k} = 10    & 2.10\%\\
  \hline
\end{tabular}
\caption{\label{err_mnist}Performance of unsupervised learning methods (without fine-tuning) with 1000 hidden units on \textbf{MNIST}.} 
}
\hspace{.1cm}
\parbox{.4\linewidth}{
\centering 
\centering 
\begin{tabular}{ |l | r| }
  \hline
  &  Error Rate \\
  \hline 
  Raw Pixels  & 23\% \\
  RBM (weight decay) & 10.6\%\\
  Dropout Autoencoder & 10.1\%\\
  Denoising Autoencoder  & 9.5\% \\
  \small{(20\% input dropout)} &\\
  \emph{k}-Sparse Autoencoder, \emph{k} = 200 & 10.4\% \\
  \emph{k}-Sparse Autoencoder, \emph{k} = 150 & 8.6\% \\
  \emph{k}-Sparse Autoencoder, \emph{k} = 75 & 9.5\%\\
  \hline
\end{tabular}
\caption{\label{err_norb}Performance of unsupervised learning methods (without fine-tuning) with 4000 hidden units on \textbf{NORB}.} 
}
\end{table*}

\subsection{Training of \emph{k}-Sparse Autoencoders} 

\subsubsection{Scheduling of the Sparsity Level} 
\label{sub:scheduling_of_the_sparsity_level}

When we are enforcing low sparsity levels in \emph{k}-sparse autoencoders (e.g., $\text{\emph{k}}$=15 on MNIST), one issue that might arise is that in the first few epochs, the algorithm greedily assigns individual hidden units to groups of training cases, in a manner similar to $k$-means clustering. In subsequent epochs, these hidden units will be picked and re-enforced and other hidden units will not be adjusted. That is, too much sparsity can prevent gradient back-propagation from adjusting the weights of these other `dead' hidden units. We can address this problem by scheduling the sparsity level over epochs as follows. 

Suppose we are aiming for a sparsity level of $\text{\emph{k}}=15$. Then, we start off with a large sparsity level (e.g. $\text{\emph{k}}=100$) for which the \emph{k}-sparse autoencoder can train all the hidden units. We then linearly decrease the sparsity level from $\text{\emph{k}}=100$ to $\text{\emph{k}}=15$ over the first half of the epochs. This initializes the autoencoder in a good regime, for which all of the hidden units have a significant chance of being picked. Then, we keep $\text{\emph{k}}=15$ for the second half of the epochs. With this scheduling, we can train all of the filters, even for low sparsity levels.

\subsubsection{Training Hyper-parameters} 

We optimized the model parameters using stochastic gradient descent with momentum as follows.
\begin{equation}
	\begin{aligned}
	\boldsymbol{v}_{k+1} &= m_\text{\emph{k}} \boldsymbol{v}_\text{\emph{k}} - \eta_\text{\emph{k}} \nabla f(\boldsymbol{x}_\text{\emph{k}}) \\
	\boldsymbol{x}_{k+1} &= \boldsymbol{x}_\text{\emph{k}} + \boldsymbol{v}_\text{\emph{k}}
	\end{aligned}
\end{equation}

Here, $\boldsymbol{v_\text{\emph{k}}}$ is the velocity vector, $m_\text{\emph{k}}$ is the momentum and $\eta_\text{\emph{k}}$ is the learning rate at the \emph{k}-th iteration. We also use a Gaussian distribution with a standard deviation of $\sigma$ for initialization of the weights. We use different momentum values, learning rates and initializations based on the task and the dataset, and validation is used to select hyperparameters.
In the unsupervised MNIST task, the values were $\sigma = 0.01$ , $m_\text{\emph{k}}=0.9$ and  $\eta_\text{\emph{k}}=0.01$, for $5000$ epochs. In the supervised MNIST task, training started with $m_\text{\emph{k}} = 0.25$ and $\eta_\text{\emph{k}} = 1$, and then the learning rate was linearly decreased to $0.001$ over $200$ epochs. In the unsupervised NORB task, the values were $\sigma = 0.01$, $m_\text{\emph{k}}=0.9$ and $\eta_\text{\emph{k}}=0.0001$, for $5000$ epochs. In the supervised NORB task, training started with $m_\text{\emph{k}} = 0.9$ and $\eta_\text{\emph{k}} = 0.01$, and then the learning rate was linearly decreased to $0.001$ over $200$ epochs.

\subsubsection{Implementations} 

While most of the conventional sparse coding algorithms require complex matrix operations such as matrix inversion or SVD decomposition, the \emph{k}-sparse autoencoders only need matrix multiplications and sorting operations in both dictionary learning stage and the sparse encoding stage. (For a parallel, distributed implementation, the sorting operation can be replaced by a method that recursively applies a threshold until $k$ values remain.) We used an efficient GPU implementation obtained using the publicly available \emph{gnumpy} library \cite{gnumpy} on a single Nvidia GTX 680 GPU.

\subsection{Effect of Sparsity Level} 
\label{sub:effect_of_sparsity_level}

In \emph{k}-sparse autoencoders, we are able to tune the value of \emph{k} to obtain the desirable sparsity level which makes the algorithm suitable for a wide variety of datasets. For example, one application could be pre-training a shallow or deep discriminative neural network. For large values of \emph{k} (e.g., $\text{\emph{k}}=100$ on MNIST), the algorithm tends to learn very local features as is shown in Figure \ref{fig_mnist}a and \ref{fig_norb}a. These features are too primitive to be used for classification using a shallow architecture since a naive linear classifier does not have enough capacity to combine these features and achieve a good classification rate. However, these features could be used for pre-training deep neural nets. 

As we decrease the the sparsity level (e.g., $\text{\emph{k}}=40$ on MNIST), the output is  reconstructed using a smaller number of hidden units and thus the features tend to be more global, as can be seen in Figure \ref{fig_mnist}b,\ref{fig_mnist}c and \ref{fig_norb}b. For example, in the MNIST dataset, the lengths of the strokes increase when the sparsity level is decreased. These less local features are suitable for classification using a shallow architecture.
Nevertheless, forcing too much sparsity (e.g., $\text{\emph{k}}=10$ on MNIST), results in features that are too global and do not factor the input into parts, as depicted Figure \ref{fig_mnist}d and \ref{fig_norb}c.  

\begin{figure}[h]
\begin{center}
\includegraphics[scale=.1]{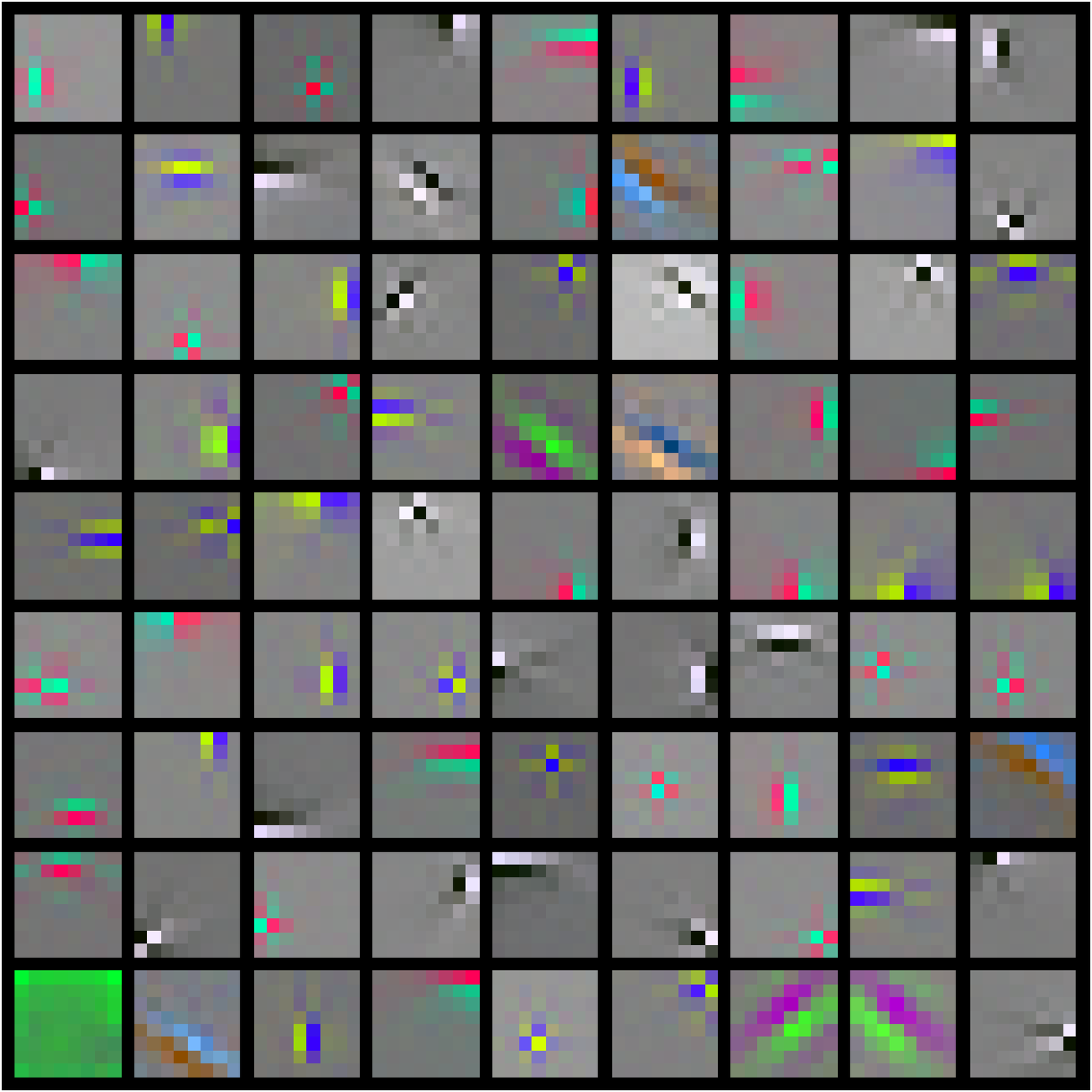}
\caption{\label{cifar10}Filters of \emph{k}-sparse autoencoder with 1000 hidden units and $\text{\emph{k}}=50$, learnt from CIFAR-10 random patches.}
\end{center}
\end{figure}

Fig. \ref{cifar10} shows the visualization of filters of the \emph{k}-sparse autoencoder with $1000$ hidden units and sparsity level of $\text{\emph{k}}=50$ learnt from random image patches extracted from CIFAR-10 dataset. We can see that the \emph{k}-sparse autoencoder has learnt localized Gabor filters from natural image patches.

\begin{figure}[h]
\begin{center}
\includegraphics[scale=.4]{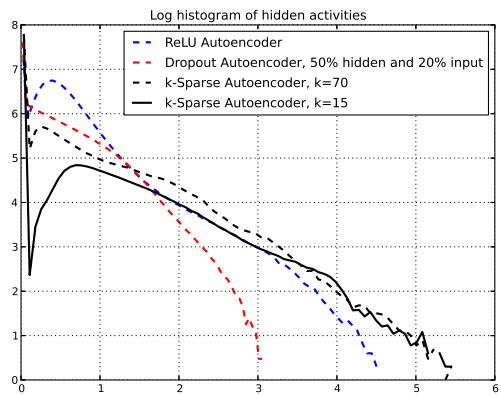}
\caption{\label{hist}Histogram of hidden unit activities for various unsupervised learning methods.}
\end{center}
\end{figure}

Fig. \ref{hist} plots histograms of the hidden unit activities for various unsupervised learning algorithms, including the \emph{k}-sparse autoencoder (\emph{k}=70 and \emph{k}=15), applied to the MNIST data. This figure contrasts the sparsity achieved by the \emph{k}-sparse autoencoder with that of other algorithms.

\subsection{Unsupervised Feature Learning Results}  
\label{sub:unsupervised_feature_learning}

In order to compare the quality of the features learnt by our algorithm with those learnt by other unsupervised learning methods, we first extracted features using each unsupervised learning algorithm. Then we fixed the features and trained a logistic regression classifier using those features. The usefulness of the features is then evaluated by examining the error rate of the classifier. 

We trained a number of architectures on the MNIST and NORB datasets, including RBM, dropout autoencoder and denoising autoencoder. In dropout, after finding the features using dropout regularization with a dropout rate of $50\%$, we used all of the hidden units as the features (this worked best). For the denoising autoencoder, after training the network by dropping the input pixels with a rate of $20\%$, we used all of the uncorrupted input pixels to find the features for classification (this worked best). In the \emph{k}-sparse autoencoder, after training the dictionary, we used $\boldsymbol{h} = \text{supp}_{\alpha\emph{k}}(W^\top  \boldsymbol{x} + \boldsymbol{b})$ to find the features as explained in Section \ref{k_sparse}, where $\alpha$ was determined using validation data. Results for different architectures are compared in Tables \ref{err_mnist}, \ref{err_norb}. We can see that the performance of our \emph{k}-sparse autoencoder is better than the rest of the algorithms. In our algorithm, the best result is achieved by $\emph{k}=25 , \alpha= 3$ with 1000 hidden units on MNIST dataset and by $\emph{k}=150 , \alpha= 2$ with 4000 hidden units on NORB dataset.

\begin{table*}[h!tb]
\parbox{.50\linewidth}{
\centering 
\begin{tabular}{ |l  | r| }
  \hline
   & Error\\
  \hline 
  Without Pre-Training  & 1.60\%\\
  RBM + F.T. & 1.24\%\\
  Shallow Dropout AE + F.T.  & 1.05\%\\
  \small{(\%50 hidden)}&\\
  Denoising AE + F.T.   & 1.20\% \\
  \small{(\%20 input dropout)}&\\
  Deep Dropout AE + F.T.  & 0.85\%\\
  \small{(Layer-wise pre-training, \%50 hidden)}&\\
  \emph{k}-Sparse AE + F.T. & 1.08\%\\
  \small{(\emph{k}=25)}&\\
  Deep \emph{k}-Sparse AE + F.T.  & 0.97\%\\
  \small{(Layer-wise pre-training)}&\\
  \hline
\end{tabular}
\caption{\label{fn_mnist}Performance of supervised learning methods on \textbf{MNIST}.  Pre-training was performed using the corresponding unsupervised learning algorithm with 1000 hidden units, and then the model was fine-tuned.}
}
\hspace{.5cm}
\parbox{.4\linewidth}{
\centering 
\begin{tabular}{ |l  | r| }
  \hline
   & Error\\
  \hline 
  Without Pre-Training  & 12.7\%\\
  DBN & 8.3\%\\
  DBM & 7.2\%\\
  third-order RBM & 6.5\%\\
  Shallow Dropout AE + F.T.  & 8.2\%\\
  \small{(\%50 hidden)}&\\
  Shallow Denoising AE + F.T.   & 7.9\%\\
  \small{(\%20 input dropout)}&\\
  Deep Dropout AE + F.T. & 7.0\%\\
  \small{(Layer-wise pre-training, \%50 hidden)}&\\
  Shallow \emph{k}-Sparse AE + F.T.  & 7.8\%\\
  \small{(\emph{k}=150)}&\\
  Deep \emph{k}-Sparse AE + F.T.  & 7.4\%\\
  \small{(\emph{k}=150, Layer-wise pre-training)}&\\
  \hline
\end{tabular}
\caption{\label{fn_norb}Performance of supervised learning methods on \textbf{NORB}. Pre-training was performed using the corresponding unsupervised learning algorithm with 4000 hidden units, and then the model was fine-tuned.} 
}
\end{table*}

\subsection{Shallow Supervised Learning Results} 
\label{sub:supervised_learning}

In supervised learning, it is a common practice to use the encoder weights learnt by an unsupervised learning method to initialize the early layers of a multilayer discriminative model \cite{pretraining}. The back-propagation algorithm is then used to adjust the weights of the last hidden layer and also to fine-tune the weights in the previous layers. This procedure is often referred to as discriminative fine-tuning. In this section, we report results using unsupervised learning algorithms such as RBMs, DBNs \cite{dbn}, DBMs \cite{dbn}, third-order RBM \cite{lifetime}, dropout autoencoders, denoising autoencoders and \emph{k}-sparse autoencoders to initialize a shallow discriminative neural network for the MNIST and NORB datasets. We used back-propagation to fine-tune the weights. The regularization method used in the fine-tuning stage of different algorithms is the same as the one used in the training of the corresponding unsupervised learning task. For instance, we fine-tuned the weights obtained from dropout autoencoder with dropout regularization or in denoising autoencoder, we fine-tuned the discriminative neural net by adding noise to the input. In a similar manner, in the fine-tuning stage of the \emph{k}-sparse autoencoder, we used the $\alpha \emph{k}$ largest hidden units in the corresponding discriminative neural network, as explained in Section \ref{k_sparse}. Tables \ref{fn_mnist} and \ref{fn_norb} reports the error rates obtained by different methods.

\subsection{Deep Supervised Learning Results} 
\label{sub:deep_supervised_learning}

The \emph{k}-sparse autoencoder can be used as a building block of a deep neural network, using greedy layer-wise pre-training \cite{layer_wise}. We first train a shallow \emph{k}-sparse autoencoder and obtain the hidden codes. We then fix the features and train another \emph{k}-sparse autoencoder on top of them to obtain another set of hidden codes. Then we use the parameters of these autoencoders to initialize a discriminative neural network with two hidden layers. 

In the fine-tuning stage of the deep neural net, we first fix the parameters of the first and second layers and train a softmax classifier on top of the second layer. We then hold the weights of the first layer fixed and train the second layer and softmax jointly using the initialization of the softmax that we found in the previous step. Finally, we jointly fine-tune all of the layers with the previous initialization. We have observed that this method of layer-wise fine-tuning can improve the classification performance compared to the case where we fine-tune all the layers at the same time.

In all of the fine-tuning steps, we keep the $\alpha \emph{k}$ largest hidden codes, where $\emph{k}=25,\alpha=3$ in MNIST and $\emph{k}=150,\alpha=2$ in NORB in both hidden layers. Tables \ref{fn_mnist} and \ref{fn_norb} report the classification results of different deep supervised learning methods.

\vspace{-.3cm}
\section{Conclusion} 
\label{sec:conclusion}

In this work, we proposed a very fast sparse coding method called \emph{k}-sparse autoencoder, which achieves exact sparsity in the hidden representation. The main message of this paper is that we can use the resulting representations to achieve state-of-the-art classification results, solely by enforcing sparsity in the hidden units and without using any other nonlinearity or regularization. We also discussed how the \emph{k}-sparse autoencoder could be used for pre-training shallow and deep supervised architectures.

\vspace{-.3cm}

\section{Acknowledgment} 
\label{sec:ack}
We would like to thank Andrew Delong, Babak Alipanahi and Lei Jimmy Ba for the valuable comments.
\bibliography{iclr}

\begin{thebibliography}{21}
\providecommand{\natexlab}[1]{#1}
\providecommand{\url}[1]{\texttt{#1}}
\expandafter\ifx\csname urlstyle\endcsname\relax
  \providecommand{\doi}[1]{doi: #1}\else
  \providecommand{\doi}{doi: \begingroup \urlstyle{rm}\Url}\fi

\bibitem[Aharon et~al.(2005)Aharon, Elad, and Bruckstein]{ksvd}
Aharon, Michal, Elad, Michael, and Bruckstein, Alfred.
\newblock K-svd: Design of dictionaries for sparse representation.
\newblock \emph{Proceedings of SPARS}, 5:\penalty0 9--12, 2005.

\bibitem[Bengio et~al.(2007)Bengio, Lamblin, Popovici, and
  Larochelle]{layer_wise}
Bengio, Yoshua, Lamblin, Pascal, Popovici, Dan, and Larochelle, Hugo.
\newblock Greedy layer-wise training of deep networks.
\newblock \emph{Advances in neural information processing systems},
  19:\penalty0 153, 2007.

\bibitem[Blumensath \& Davies(2009)Blumensath and Davies]{iht}
Blumensath, Thomas and Davies, Mike~E.
\newblock Iterative hard thresholding for compressed sensing.
\newblock \emph{Applied and Computational Harmonic Analysis}, 27\penalty0
  (3):\penalty0 265--274, 2009.

\bibitem[Coates \& Ng(2011)Coates and Ng]{adam}
Coates, Adam and Ng, Andrew.
\newblock The importance of encoding versus training with sparse coding and
  vector quantization.
\newblock In \emph{Proceedings of the 28th International Conference on Machine
  Learning (ICML-11)}, pp.\  921--928, 2011.

\bibitem[Coates et~al.(2011)Coates, Ng, and Lee]{cifar10_adam}
Coates, Adam, Ng, Andrew~Y, and Lee, Honglak.
\newblock An analysis of single-layer networks in unsupervised feature
  learning.
\newblock In \emph{International Conference on Artificial Intelligence and
  Statistics}, pp.\  215--223, 2011.

\bibitem[Donoho \& Elad(2003)Donoho and Elad]{donoho}
Donoho, David~L and Elad, Michael.
\newblock Optimally sparse representation in general (nonorthogonal)
  dictionaries via ℓ1 minimization.
\newblock \emph{Proceedings of the National Academy of Sciences}, 100\penalty0
  (5):\penalty0 2197--2202, 2003.

\bibitem[Engan et~al.(1999)Engan, Aase, and Hakon~Husoy]{mod}
Engan, Kjersti, Aase, Sven~Ole, and Hakon~Husoy, J.
\newblock Method of optimal directions for frame design.
\newblock In \emph{Acoustics, Speech, and Signal Processing, 1999.
  Proceedings., 1999 IEEE International Conference on}, volume~5, pp.\
  2443--2446. IEEE, 1999.

\bibitem[Erhan et~al.(2010)Erhan, Bengio, Courville, Manzagol, Vincent, and
  Bengio]{pretraining}
Erhan, Dumitru, Bengio, Yoshua, Courville, Aaron, Manzagol, Pierre-Antoine,
  Vincent, Pascal, and Bengio, Samy.
\newblock Why does unsupervised pre-training help deep learning?
\newblock \emph{The Journal of Machine Learning Research}, 11:\penalty0
  625--660, 2010.

\bibitem[Gregor \& LeCun(2010)Gregor and LeCun]{fast}
Gregor, Karol and LeCun, Yann.
\newblock Learning fast approximations of sparse coding.
\newblock In \emph{Proceedings of the 27th International Conference on Machine
  Learning (ICML-10)}, pp.\  399--406, 2010.

\bibitem[Hinton et~al.(2012)Hinton, Srivastava, Krizhevsky, Sutskever, and
  Salakhutdinov]{dropout}
Hinton, Geoffrey~E, Srivastava, Nitish, Krizhevsky, Alex, Sutskever, Ilya, and
  Salakhutdinov, Ruslan~R.
\newblock Improving neural networks by preventing co-adaptation of feature
  detectors.
\newblock \emph{arXiv preprint arXiv:1207.0580}, 2012.

\bibitem[Kavukcuoglu et~al.(2010)Kavukcuoglu, Ranzato, and LeCun]{psd}
Kavukcuoglu, Koray, Ranzato, Marc'Aurelio, and LeCun, Yann.
\newblock Fast inference in sparse coding algorithms with applications to
  object recognition.
\newblock \emph{arXiv preprint arXiv:1010.3467}, 2010.

\bibitem[LeCun et~al.(2004)LeCun, Huang, and Bottou]{norb}
LeCun, Yann, Huang, Fu~Jie, and Bottou, Leon.
\newblock Learning methods for generic object recognition with invariance to
  pose and lighting.
\newblock In \emph{Computer Vision and Pattern Recognition, CVPR}, volume~2,
  pp.\  II--97. IEEE, 2004.

\bibitem[Lee et~al.(2007)Lee, Ekanadham, and Ng]{ng}
Lee, Honglak, Ekanadham, Chaitanya, and Ng, Andrew.
\newblock Sparse deep belief net model for visual area v2.
\newblock In \emph{Advances in neural information processing systems}, pp.\
  873--880, 2007.

\bibitem[Maleki(2009)]{coherence}
Maleki, Arian.
\newblock Coherence analysis of iterative thresholding algorithms.
\newblock In \emph{Communication, Control, and Computing, 2009. Allerton 2009.
  47th Annual Allerton Conference on}, pp.\  236--243. IEEE, 2009.

\bibitem[Nair \& Hinton(2009)Nair and Hinton]{lifetime}
Nair, Vinod and Hinton, Geoffrey~E.
\newblock 3d object recognition with deep belief nets.
\newblock In \emph{Advances in Neural Information Processing Systems}, pp.\
  1339--1347, 2009.

\bibitem[Olshausen \& Field(1997)Olshausen and Field]{sparse_coding}
Olshausen, Bruno~A and Field, David~J.
\newblock Sparse coding with an overcomplete basis set: A strategy employed by
  v1?
\newblock \emph{Vision research}, 37\penalty0 (23):\penalty0 3311--3325, 1997.

\bibitem[Salakhutdinov \& Larochelle(2010)Salakhutdinov and Larochelle]{dbn}
Salakhutdinov, Ruslan and Larochelle, Hugo.
\newblock Efficient learning of deep boltzmann machines.
\newblock In \emph{International Conference on Artificial Intelligence and
  Statistics}, pp.\  693--700, 2010.

\bibitem[Tieleman(2010)]{gnumpy}
Tieleman, Tijmen.
\newblock Gnumpy: an easy way to use gpu boards in python.
\newblock \emph{Department of Computer Science, University of Toronto}, 2010.

\bibitem[Tropp \& Gilbert(2007)Tropp and Gilbert]{omp}
Tropp, Joel~A and Gilbert, Anna~C.
\newblock Signal recovery from random measurements via orthogonal matching
  pursuit.
\newblock \emph{Information Theory, IEEE Transactions on}, 53\penalty0
  (12):\penalty0 4655--4666, 2007.

\bibitem[Van~Gemert et~al.(2008)Van~Gemert, Geusebroek, Veenman, and
  Smeulders]{mix}
Van~Gemert, Jan~C, Geusebroek, Jan-Mark, Veenman, Cor~J, and Smeulders,
  Arnold~WM.
\newblock Kernel codebooks for scene categorization.
\newblock In \emph{Computer Vision--ECCV 2008}, pp.\  696--709. Springer, 2008.

\bibitem[Vincent et~al.(2008)Vincent, Larochelle, Bengio, and Manzagol]{da}
Vincent, Pascal, Larochelle, Hugo, Bengio, Yoshua, and Manzagol,
  Pierre-Antoine.
\newblock Extracting and composing robust features with denoising autoencoders.
\newblock In \emph{Proceedings of the 25th international conference on Machine
  learning}, pp.\  1096--1103. ACM, 2008.

\end{thebibliography}
\bibliographystyle{icml2014}

\end{document}